\documentclass[11pt,a4paper]{article}

\usepackage{linguex}

\usepackage[hyperref]{emnlp2020}
\usepackage{times}
\usepackage{latexsym}

\usepackage{url}
\usepackage{graphicx}
\usepackage{tabularx}
\usepackage{multirow}
\usepackage{booktabs}
\newcolumntype{L}{>{\raggedright\arraybackslash}X}
\usepackage{pifont}

\usepackage{microtype}

\aclfinalcopy % Uncomment this line for the final submission
\def\aclpaperid{1379} %  Enter the acl Paper ID here

\title{{X-SRL}: A Parallel Cross-Lingual Semantic Role Labeling Dataset}

\author{Angel Daza$^{\dagger}$\ding{168} and Anette Frank\ding{168} \\
   Leibniz ScienceCampus ``Empirical Linguistics and Computational Language Modeling''\\
   \ding{168}Heidelberg University / $^{\dagger}$Institut f\"{u}r Deutsche Sprache Mannheim\\
   Germany\\
  {\tt \{daza,frank\}@cl.uni-heidelberg.de}
 \\}

\date{}

\begin{document}
\maketitle
\begin{abstract}
  Even though SRL is researched for many lan\-gua\-ges, major improvements have mostly been obtain\-ed for English, for which more re\-sour\-ces are available. In fact, existing multi\-lin\-gual SRL datasets contain disparate anno\-tation styles or come from different domains, hampering generalization in multi\-lin\-gu\-al learning. In this work we propose a method to auto\-ma\-tically construct an SRL corpus that is \textit{parallel in four lan\-gua\-ges: Eng\-lish, French, German, Spanish}, with unified predicate and role annotations that are fully com\-pa\-ra\-ble  across languages. We apply high-\-qua\-li\-ty machine translation to the English CoNLL-09 dataset and use multilingual BERT to project its high-quality annotations to the target languages. We include hu\-man-\-va\-li\-da\-ted test sets that we use to measure the pro\-jection quality, and show that projection is denser and more precise than a strong baseline. Finally, we train different SOTA models on our novel corpus for mono- and multilingual SRL, 
  %\af{which demonstrates the generalization capacities of our new corpus}. 
  showing that the multilingual annotations improve performance especially for the weaker languages.
\end{abstract}

\section{Introduction}
\label{intro}

Semantic Role Labeling (SRL) is the task of extracting semantic predicate-argument structures from sentences. One of the most widely used labeling schemes for this task is based on PropBank \cite{Palmer05-SRL}. It comes in two va\-ri\-ants: span-based labeling, where arguments are characterized as word-spans \cite{Carreras05-SRL,PradhanCoNLL2012}, and head-based labeling, which only labels the syntactic head \cite{Hajic09-SRL}. In this work we focus on head-based labeling, as it is applied in the multilingual CoNLL-09 shared task dataset, comprising 7 languages.

\begin{figure}
\centering
  \includegraphics[width=0.48\textwidth]{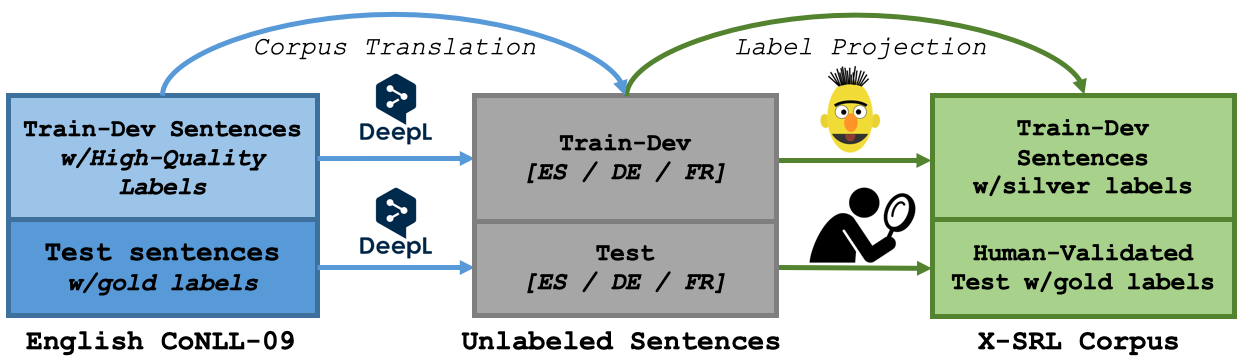}
  \caption{Method to create X-SRL. We automatically translate the English CoNLL-09 corpus, use a fast label projection method for \textit{train-dev} and get human annotators to select the appropriate head words on the target sentences to obtain gold annotations for the \textit{test} sets.}
\label{fig:method}
\end{figure}

The performance of English SRL has considerably improved in recent years through continuous refinements of Deep Neural Network (DNN) models \cite{Zhou15-SRL,He17-SRL,March17-SRL,Cai18-SRL}; however, although the CoNLL-09 SRL dataset already covers 7 languages, other languages have not received the same
level of attention. This situation may be due to factors such as i) the lack of sufficient training data to successfully apply a language-agnostic DNN model; ii) the fact that creating new SRL datasets is resource-consuming; iii) current label projection methods suffering from low recall; finally, iv) even in cases where annotated resources are available in other languages, often they were automatically converted from independent pre-existing annotation schemes or labeled with automatic methods, resulting in data quality and labeling schema divergences, hampering the effectiveness of unified models that can be applied in multilingual settings. 

In this paper we offer a \textbf{multilingual parallel SRL corpus -- X-SRL --} for English (EN), German (DE), French (FR) and Spanish (ES) that is based on English gold annotations and \textit{shares the same labeling scheme across languages}.\footnote{\url{https://github.com/Heidelberg-NLP/xsrl_mbert_aligner}} Our corpus has two major advantages compared to existing datasets: first, since it is a \textit{parallel corpus}, all sentences are semantically equivalent allowing us to analyze performance at the sentence-level and to better understand the reasons for SRL score divergences across languages\footnote{E.g., German F1 score on the CoNLL-09 dataset lags 10 points behind English, but with currently available datasets we cannot be sure if this is due to differences in the available training data or because of language-specific characteristics.}; second, we expect that models trained jointly on multiple languages annotated with a homogeneous labeling style will be able to better generalize across languages\footnote{It is not straightforward to use the CoNLL-09 data in a multilingual model: for example, annotations for German use a role inventory with roles A0-A9, and a one-to-one mapping to all English labels is not available.}. Moreover, by minimizing the need of specialized human annotators, our parallel corpus construction method is \textit{lightweight and portable}, since it is built on three main components: i) A high-quality annotated dataset in the source language, in this case the English CoNLL-09 corpus \cite{Hajic09-SRL}, ii) a high-quality SOTA Machine Translation system, we are using DeepL Pro\footnote{\url{https://www.deepl.com/translator}}; and iii) multilingual contextual word representations, in this case multilingual BERT (mBERT) \cite{Devlin18-bert-exps}.  The situation for these multilingual resources is improving with each day, and thus our method, in perspective, could be followed for producing training data for more lower-resource languages.
Importantly, although we automatically project labels from English to the newly available corpora in the different languages for \textit{train} and \textit{dev} sections, we also provide \textit{test} sets on which humans assess the quality of the automatic translations and select the valid predicates as well as the appropriate target head words for the target sentences. Having a human-validated test set ensures solid benchmarking for SRL systems, and additionally allows us to assess the validity of the proposed automatic projection for the rest of the data.

Our projection method works as follows (see also Figure \ref{fig:method}):
We obtain automatic translations of the English CoNLL-09 corpus into each of our target languages; then, we automatically label them without applying any language-pair specific label projection model, but use mBERT with additional filters as a means for alignment. We show that by following this approach we obtain a \textit{more densely annotated dataset} compared to an existing SOTA label projection method \cite{akbik18-zap}. In short, our contributions are:

\begin{itemize}
    \item The first fully parallel SRL dataset with dense, homogeneous annotations and human-validated test sets covering four languages: English, French, German and Spanish.
    \item A simple but effective novel method to project existing SRL annotations from English to lower-resource languages.
    \item A fast method to create a human-supervised test set that allows us to explore the syntactic and semantic divergences in SRL across languages and to assess performance differences. 
    \item We provide quality measures for our projection based on the human validation process.
    \item We demonstrate the multilingual generalization capabilities of our corpus by training different SOTA baseline models on our dataset.
\end{itemize}
\section{Related Work}
\label{rel_work}

Semi-automatic annotation projection has been applied to different SRL frameworks. \citet{Pado-07, Pado09-Proj} proposed a projection method for FrameNet \cite{Baker98-SRL} semantic roles that searches for the best alignment of source and target constituent trees, and also created a small human-validated test set for benchmarking.

A number of PropBank resources are available for different languages: the benchmark datasets CoNLL-09 \cite{Hajic09-SRL} and CoNLL-12 \cite{PradhanCoNLL2012} are well-established,  however,  a direct cross-lingual comparison of SRL performance across the covered languages is not possible. The reason being that the language-specific datasets come from different sources and were not conceived for such a comparison. 

On the other hand, \citet{vdPlas10-Proj} attested the validity of English PropBank labels for French and directly applied them on French data. This motivated SRL projection methods such as \citet{vanDerPlas11-Proj} and \citet{Akbik15-Proj}, which aim to generate a common label set across languages. A known issue with this approach is the need for good quality parallel and sentence-level filtered data. For this reason they used existing parallel corpora, Europarl \cite{Koehn05-Euro} and UN \cite{Ziemski16-UN}, automatically labeled the English side with SRL annotations and transferred them to the corresponding translations. The major issue with this is that evaluation against ground-truth and detailed error analysis on the target languages are not possible, since all annotations are automatic and come from noisy sources. Likewise, the Universal Proposition Bank\footnote{\url{https://github.com/System-T/UniversalPropositions}}\citep{Akbik15-Proj, Akbik-PolyglotSRL}, adds an SRL layer on top of the Universal Dependencies \cite{de-marneffe-14-universalDep} corpus, which covers eight different languages. However, i) the original corpora come from independent sources and are not parallel, ii) the source sentences were automatically labeled containing noise even before the alignment step, and iii) the test sets also contain automatic projections without human validation of the labels.

In contrast, we present a corpus that transfers English high-quality labels to the target side, thus projecting the same labeling style to the other languages; more importantly, we conceive of this corpus, from the very beginning, as a parallel resource with translation equivalence with the source and target languages at the sentence-level. In addition, we create a human-validated test set to allow for proper benchmarking in each language. 

The use of synthetic data generated by automatic translation has proven to improve performance for MT \cite{sennrich-etal-2016-improving} and Argumentation Mining \cite{eger-18-cross-AM}. We similarly create a parallel corpus using automatic translation, however, to our knowledge, we are the first to create a directly comparable multilingual SRL dataset using automatic translation with a manually validated test set, minimizing human labour. 

Another attempt to close the gap between languages is by training multilingual SRL systems. \citet{he-19-syntax} propose a biaffine scorer with syntax rules to prune of candidates, achieving SOTA independently in all languages from CoNLL-09. \citet{Mulcaire18-SRL} and \citet{daza-frank-2019-translate} train a single model using input data from different languages and obtain modest improvements, especially for languages where less monolingual training data is available. In this sense, our X-SRL corpus contributes with more compatible training data across languages, and aims to improve the performance of jointly trained multilingual models. 

% \angel{\textbf{mBERT.} Multilingual BERT \cite{Devlin18-bert-exps} is a deep contextualized language model based on the transformer architecture \cite{Vaswani-17-trans} that was trained with data from Wikipedia in 104 languages. The model was not explicitly trained with a cross-lingual signal, however it implements a \textit{universal tokenization} that generates shared word pieces across languages, contextualizing similar words in a shared space regardless of their language of origin. It has been widely shown that these pre-trained language models are suitable when fine-tuning them for several NLP tasks, including token-level classification, and particularly mBERT has shown promise when trained with one language and zero-shot tested on other languages \cite{wu-dredze-19-beto}.}

\section{Building X-SRL}
\label{dataset}

In this section we first explain our method for translating the English CoNLL-09 SRL dataset (\S 3.1) into three target languages (DE, ES, FR) \footnote{We chose these languages given the availability of annotators to validate the quality of test set translations. We hope that future work will apply our method to further languages.}. In \S 3.2 we describe how the human-validated labels (only for the test sets) were obtained in an efficient way, and report annotator agreement statistics in \S 3.3. The details of how we perform (automatic) label projection enhanced with filtering for \textit{train/dev} are given in \S 3.4. With this we achieve big annotated SRL datasets for three new languages (cf.\ Table \ref{tab:X-SRL_Data}).

When building the X-SRL dataset, in line with the current PropBank SRL data available in different languages, we focus on verbal predicates only. 
Note that the English CoNLL-09 data includes both verbal and nominal predicate annotations, yet this is due to the NomBank project \cite{meyers-04-nombank} being available for that language. By contrast, the remaining languages with PropBank SRL training data (including the CoNLL-09 non-English data) only provide annotations for verbal predicates. While we could attempt projecting the English nominal predicate annotations and create an X-SRL dataset that includes nominal SRL for all target languages -- which would mean a big advance over the current situation -- admitting nominal and verbal SRL annotations in a multilingual setting would confront us with many translation shifts. We could try to capture these for the manually curated test set, however we would run a risk of generating noisy or scarce target annotations when projecting them for the \textit{train/dev} sections.\footnote{The reasons are complex: First, by including nominal SRL, we would be confronted with translation shifts in both directions, e.g. N-to-V or V-to-N translations. For these, we'd have to verify whether they correspond to valid verbalizations or nominalizations on the target side. This would lead to considerable overhead and, most likely, noise in automatic projection. Also, translation shifts often involve light verb constructions, which require special role annotations. These would be difficult to assign in automatic projection. We thus defer the inclusion of nominal SRL to future work. \label{fn:nominalSRL}}

\begin{table*}
% \begin{tabularx}{\linewidth}{cL}
% \toprule
\vspace*{.5mm}

\scriptsize

\ex.  \a.  People aren't \textbf{panicking}.                  
      \bg. La gente no está \textbf{entrando} \textbf{en} \textbf{pánico}.\\
           The people not are entered in panic.\\
\vspace*{.3mm}

% \midrule

\ex. \a. The account had \textbf{billed} about \$6 million in 1988, according to Leading National Advertisers.                               \bg. Das Konto hatte 1988 etwa 6 Millionen Dollar \textbf{in} \textbf{Rechnung} \textbf{gestellt}, so die Leading National Advertisers.\\
The account had 1988 about 6 million dollars in invoice put, so the Leading National Advertisers.\\
\vspace*{.3mm}

% \midrule

\ex. \a. The economy does, however , \textbf{depend} on the confidence of businesses, consumers and foreign investors .   
    \bg. Die Wirtschaft \textbf{h\"angt} jedoch vom Vertrauen von Unternehmen, Verbrauchern und ausl\"andischen Investoren \textbf{ab}. \\ 
     The economy hangs however from-the confidence of businesses, consumers and foreign investors off .\\
\vspace*{.3mm}

% \midrule  

\ex. \a. But while the \textbf{New} \textbf{York} \textbf{Stock} \textbf{Exchange} did n't fall apart Friday as the \textbf{Dow} \textbf{Jones} \textbf{Industrial} \textbf{Average} plunged 190.58 points. 
     \bg. Mais si la \textbf{Bourse} \textbf{de} \textbf{New} \textbf{York} ne s' est pas effondrée vendredi alors que le \textbf{Dow} \textbf{Jones} \textbf{Industrial} \textbf{Average} a chuté de 190,58 points.\\
     But if the Exchange of New York not Refl is not collapsed Friday when that the Dow Jones Industrial Average has fallen by 190.58 points.\\
     \vspace*{.3mm}

% \bottomrule 

% \end{tabularx}

\caption{\label{tab:special_cases}
Examples of translation shifts:  (1) predicate nominalization on the target side, (2) and (1) source verb converted to a light verb construction on the target side, (3) a source predicate translates to a verb with separable prefix, and (4) instances of Named Entities being translated or not to the target language.}

\end{table*}

\subsection{Dataset Translation}

We aim to produce high-quality labeled corpora while reducing as much as possible the amount of time, cost and human intervention needed to fulfill this task. We use Machine Translation to perform dataset translation, obviating the need of human translator services. As previous work \cite{Tiedemann-16-AnnoMT, tyers-18-AnnoMT} has shown, automatic translations are useful as supervision for syntactic dependency labeling tasks since they are quite close to the source languages; likewise, in Argumentation Mining, \citet{eger-18-cross-AM} achieve comparable results to using human-translated data. One could argue that by automatically translating the English source, we could run into a problem of  \textit{translationese}.\footnote{Translationese occurs when -- in an attempt to reproduce the meaning of a text in a foreign language -- the resulting translation is grammatically correct but carries over language-specific constructs from the source language to the target} While it would be interesting to study possible shining-through effects in our automatically translated target texts and any potential impact on SRL performance (e.g. by comparing a natural vs.\ translated test set), our main concern is to preserve the relevant \textit{predicate-argument structures} in order to give a strong-enough signal to train our SRL systems, and our initial assumption relies on the evidence from the mentioned previous works (confirmed by our results) that obtaining relevant training data is possible with MT generated sentences.

We take as source the set of sentences in the English CoNLL-09 dataset, which are tokenized and annotated for part-of-speech (POS), syntactic dependencies, predicate senses and semantic roles. We use DeepL to obtain translations of each sentence into the three target languages. For all target sentences we use spaCy\footnote{\url{https://github.com/explosion/spaCy}} to tokenize, assign POS tags and syntactic dependency annotations. This gives us a 4-way parallel corpus with syntactic information on both sides.

\subsection{Test Set Annotation}

\textbf{Annotation Setup.} To confirm the good quality of the translations delivered by DeepL, we hired 12 annotators with a background in translation studies and experience in $EN\rightarrow T$ translation (we hired 4 annotators for each language pair) to \textit{rate and validate the automatic translations of the test set}\footnote{Note that validating a translation that already exists is considerably faster than generating translations from scratch, therefore annotation time and budget dropped significantly.} by following a guideline that explains the quality validation and the annotation processes\footnote{See Supplement A for the annotation guideline.}. First, we ask them to rate the translations on a scale from 1-5 (worst to best). On the basis of the obtained ratings, we apply a filter and keep only the sentences with \textit{quality rating 3, 4, or 5}. Only on this subset of sentences we require them to do three more tasks: i) we show them the \textit{labeled verbal predicates}\footnote{We ignore all source nominal predicates.} in the English sentence and ask them to mark on the target side the words that express the same meaning, ii) we show them a list of \textit{key arguments} (which correspond to the labeled syntactic heads in the English sentence) and likewise, ask them to mark on the target side the expression that best matches each key argument's meaning (marking several words is allowed), and finally iii) we ask them to \textit{fix minor translation mistakes} in order to better reflect the source meaning. Importantly, we ask annotators to flag as \textbf{special cases} any one-to-many mappings, and for predicates, any mapping that aligns a source verb to a non-verbal predicate in the target language. We also give the option to map source heads or predicate words to \textit{NONE} when no relevant corresponding expression in the translated sentence can be found.

\textbf{Annotation Agreement.}
To approximate the inter-annotator agreement, we gave the first 100 sentences to all annotators of each language pair and compute Krippendorff's alpha\footnote{We use the NLTK implementation with binary distance to compute the agreement of labels.} on this sub-set of sentences. We obtain $\alpha_{pred_{DE}}$=0.75, $\alpha_{pred_{ES}}$=0.73, $\alpha_{pred_{FR}}$=0.78 for \textit{predicate} and $\alpha_{role_{DE}}$=0.79, $\alpha_{role_{ES}}$=0.70, $\alpha_{role_{FR}}$=0.79 for \textit{role labels}. This shows that the \textit{fast} annotation method can be trusted.

\textbf{Linguistic Validation.} We run a second annotation round where two annotators with linguistic background re-validate the instances that were flagged as \textbf{special cases} by translators during the first round (more concretely, the possible \textit{translation shifts}). Specifically, annotators in this phase decide, for each special case, if the annotated label should be deleted or corrected.  The cases could fall into one or more of the following categories\footnote{This validation was performed independently, according to the annotators' language expertise. However, the annotators discussed general policies and jointly resolved difficult cases.} (see Table \ref{tab:special_cases} for some examples):

\begin{itemize}
    \item \textbf{Nominalizations:}  A verbal expression (predicate) in English is translated to a nominal expression in the Target (see Table \ref{tab:special_cases}, examples (1, 2)). Since we restrict our dataset to verbal predicates (see fn.\ \ref{fn:nominalSRL}) we discourage the annotation of nominal predicates even when they preserve the original sense.
     
    \item \textbf{Light Verb Constructions:} This is a special case of nominalization on the target side, where a noun that corresponds to a verb in the source language is an argument of a so-called 'light' verb with bleached, often aspectual, meaning. In example (2), the verb \textit{billed} is translated to \textit{in Rechnung gestellt} (literally: 'in invoice put'). According to \citet{bonial-12-propbank}, the nominal argument of a light verb needs a special role annotation.\footnote{The noun projects its predicate-specific role set and in addition includes the governing verb with a role ARGM-LVB.} Since there is no easy automatic method to figure out the target senses, we leave these cases for future work and do not annotate them here.
    
    \item \textbf{Separable Verb Prefixes:} In German, specific verbs must split off their prefix in certain constructions, even though this prefix crucially contributes to their meaning. In example (3), the German verb is \textit{abhängen} which means \textit{to depend}, while the verb \textit{hängen} means \textit{to hang}. Since the labeling scheme that we are using only allows us to tag one word as the head, annotators were instructed to pick the truncated stem of the verb, given that the particle is a syntactic dependent of it.
    
    \item \textbf{Multiword Expressions (MWEs):} A single source word is translated to several target words that constitute a single unit of meaning. The translators were allowed to mark more than one target word if the source word meaning could be mapped to a MWE. For these cases, if they did not fall in any of the previous three categories, and since they were manually aligned for being equivalent in meaning, we transfer the source label to the syntactic head of the marked MWE.  
    
    \item \textbf{Named Entities:} are treated as special cases of MWEs. Some NEs, but not all, are (correctly) translated to the target language, which can result in a change of the argument's head. We see both cases in example (4). When NEs are translated to the target language, we need to select the appropriate head: \textit{Exchange} is the head of the NE in English but \textit{Bourse} should be the head in French. We re-locate the label to the NE's syntactic head on the target side.

\end{itemize}

The linguistic analysis highlights the importance of providing a human-validated test set -- as opposed to relying on automatic projection. While the English labels are considered to be gold standard, their transfer to any target language is not straightforward and must be controlled for the mentioned cases to be considered gold standard on the target side. Accordingly, we also to consider filters or refinements for the automatic projection and finally, on the basis of our validated test set, we can evaluate how accurate our automatic projection is.

\begin{table}[t!]
\centering
\resizebox{0.43\textwidth}{!}{

\begin{tabular}{@{}lrrrr@{}}
\toprule
QUALITY (Q)                             & EN     & DE     & ES     & FR     \\ 
\cmidrule{1-2}  \cmidrule{3-5}
5                                   & 2,399  & 718    & 1,758  & 1,358  \\
4                                   & 0      & 902    & 407    & 463    \\
3                                   & 0      & 593    & 181    & 274    \\
2                                   & 0      & 164    & 46     & 184    \\
1                                   & 0      & 22     & 15     & 119    \\ 
\midrule
\# Sentences Q \textgreater 2       & 2,399  & 2,213  & 2,346  & 2,095  \\
\# Kept Predicates Q \textgreater 2 & 5,217  & 4,086  & 4,376  & 3,770  \\
\# Kept Arguments Q \textgreater 2  & 14,156 & 11,050 & 10,529 & 9,854 \\ \bottomrule
\end{tabular}

}
\caption{EN shows the original numbers for the English CoNLL-09 corpus. The other three languages show the quality distribution and \textit{predicate} and \textit{role} annotations kept after applying the quality and linguistic filters. 
}
\label{Test-Data-Dist}
\end{table}

% \begin{figure*}
% \centering
%   \includegraphics[width=0.62\textwidth]{figs/Matrix_Example2.png}
%   \caption{We compute a pair-wise cosine similarity matrix to simulate word alignments. For each column (source word pieces) we keep the top-k (k=2) most similar target-side word piece candidates (red squares). By mapping word pieces to their full-words and applying filters we 
%   choose the final aligned target words for each source word.
%  % align each considered full source word to the final target word.}
%  }
% \label{fig:alignments}
% \end{figure*}

\subsection{Test Statistics} 
Table \ref{Test-Data-Dist} shows the statistics for the final quality distribution for each of the target language datasets according to the translators' ratings. The final test sets are composed by all sentences with quality level higher than 2. We observe that after applying this filter, the three languages have roughly similar amounts of good quality sentences (between 87\% and 97\%) as well as similar density of annotations for both predicate and argument labels. The number of sentences that are completely 4-way parallel is 1,714 (71.45\% of the original EN corpus). This confirms the intuition that DeepL generates translations that are faithful to the sources. The number of \textit{special cases} analyzed in the second validation step were 294 (DE), 332 (ES) and 1300 (FR), of which 105, 122 and 173, respectively, were considered to be translation shifts and thus were not considered further.

\subsection{Automatic Projection}

The next step is to find an efficient method to automatically transfer the labels in the \textit{train/dev} portions of the data to the target languages without loosing too many gold labels. Contrary to the test set, we cannot perform human validation due to the size of the data; here we are mostly interested in getting automatically \textit{good enough} labels to train models. Usually, label projection methods \cite{Pado-07,Pado09-Proj, vanDerPlas11-Proj, Akbik15-Proj,AminianRD19-RawSRL} rely on the intersection of \textit{source-to-target} and \textit{target-to-source} word alignments to transfer the labels in the least noisy manner, and this way prefer to have higher precision at the expense of lower recall. Instead, we take a novel approach and rely on the shared space of mBERT embeddings \cite{Devlin18-bert-exps}. Specifically, we compute pair-wise cosine similarity between source and target tokens and emulate word-alignments according to this measure\footnote{This is similar to what is done as a first step in BERTScore \cite{Zhang20-bertScore} towards computing a metric for (semantic) sentence similarity, but here we use the token-wise similarity as a guide for cross-lingual word alignments.}. We show that using mBERT instead of typical word alignments dramatically improves the recall of the projected annotations, and enhanced with filters, it also achieves high enough precision, resulting in a more densely labeled target side and therefore better quality training data is expected. Additionally, previous works show that BERT contextualized representations are useful for monolingual Word Sense Disambiguation (WSD) tasks 
\cite{loureiro-19-BERT-WSD,huang-19-glossbertWSD} which lets us assume that we can rely on mBERT to find good word-level alignments across languages.

\begin{figure}[t]
\centering
  \includegraphics[width=0.45\textwidth]{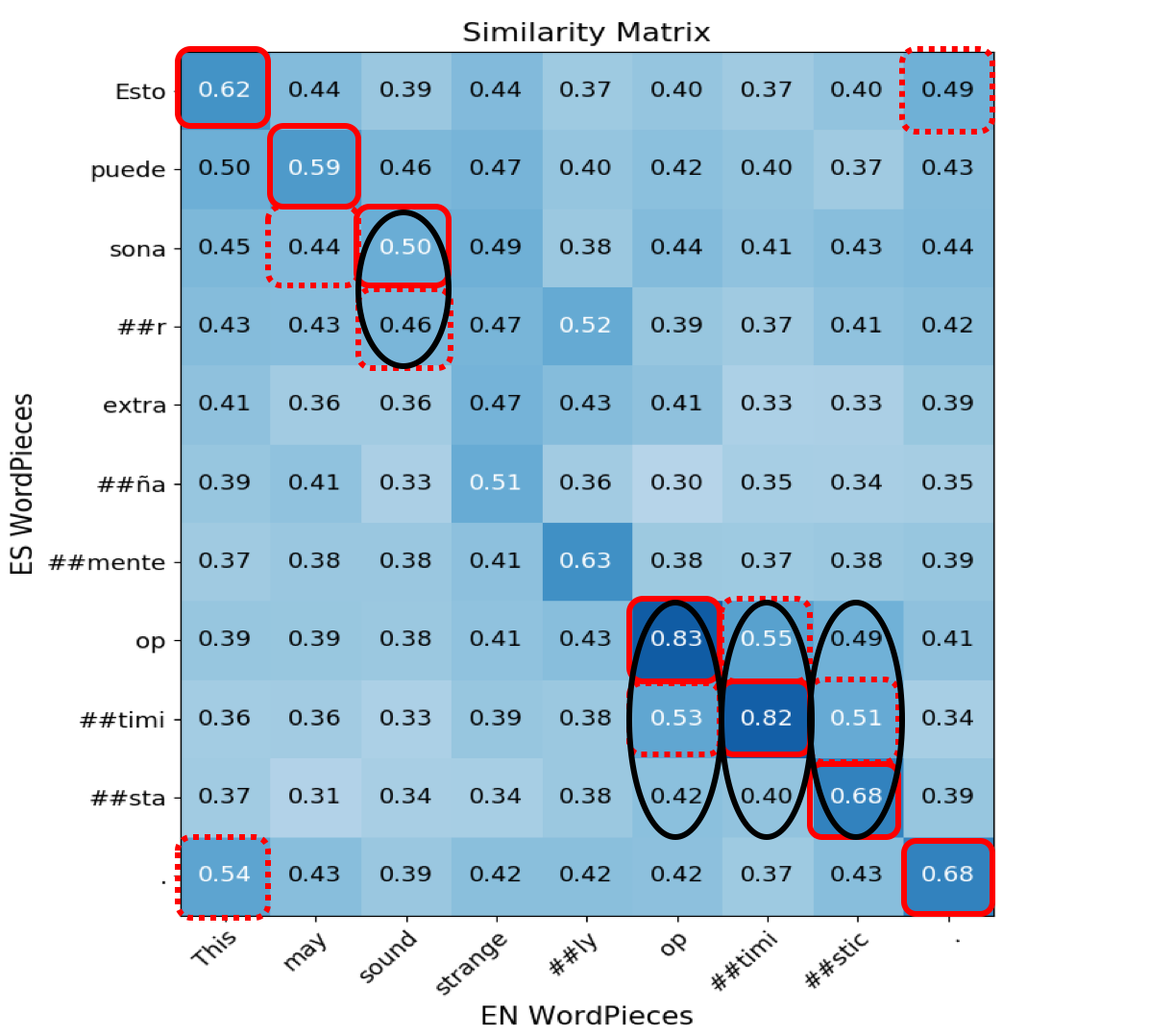}
  \caption{We compute a pair-wise cosine similarity matrix to simulate word alignments. For each column, we look only at source word-pieces with an associated label and keep the top-k (k=2) most similar target-side word piece candidates (red squares). The black circles show the aligned full-word. 
 % By mapping word pieces to their full-words and applying filters we choose the final aligned target words for each source word.\\
}
\label{fig:alignments_matrix}
\end{figure}

\textbf{BERT Cosine Similarity.} 
We start with our word tokenized parallel source $\mathbf{S}=(w_{s_0}, ..., w_{s_n})$ and target $\mathbf{T}=(w_{t_0}, ..., w_{t_m})$ sentences. Then, we use the mBERT tokenizer to obtain word-pieces and their corresponding vectors $\mathbf{S'}=(v_{s_0}, ..., v_{s_p})$ and $\mathbf{T'}=(v_{t_0}, ..., v_{t_q})$ respectively, where we have $p$ source word-pieces and $q$ target word-pieces. We compute the pairwise word-piece cosine similarity between $\mathbf{S'}$ and $\mathbf{T'}$. The cosine similarity between a source word-piece vector and a target word-piece vector is $\frac{ v_{s}^{T}v_{t} }{\left | \left | v_{s}\right | \right | \left | \left |v_{t}\right | \right |}$ \footnote{We use the implementation of \citet{Zhang20-bertScore}.}. The result is a similarity matrix $\mathbf{SM}$ with $p$ (columns) and $q$ (rows) word-pieces (see Figure \ref{fig:alignments_matrix}). In addition, we keep a mapping $\mathbf{S'}\rightarrow \mathbf{S}$ and $\mathbf{T'}\rightarrow \mathbf{T}$ from each of the word-piece vectors to their original respective word tokens to recover the full-word alignments when needed.

\begin{figure}[t]
\centering
  \includegraphics[width=0.45\textwidth]{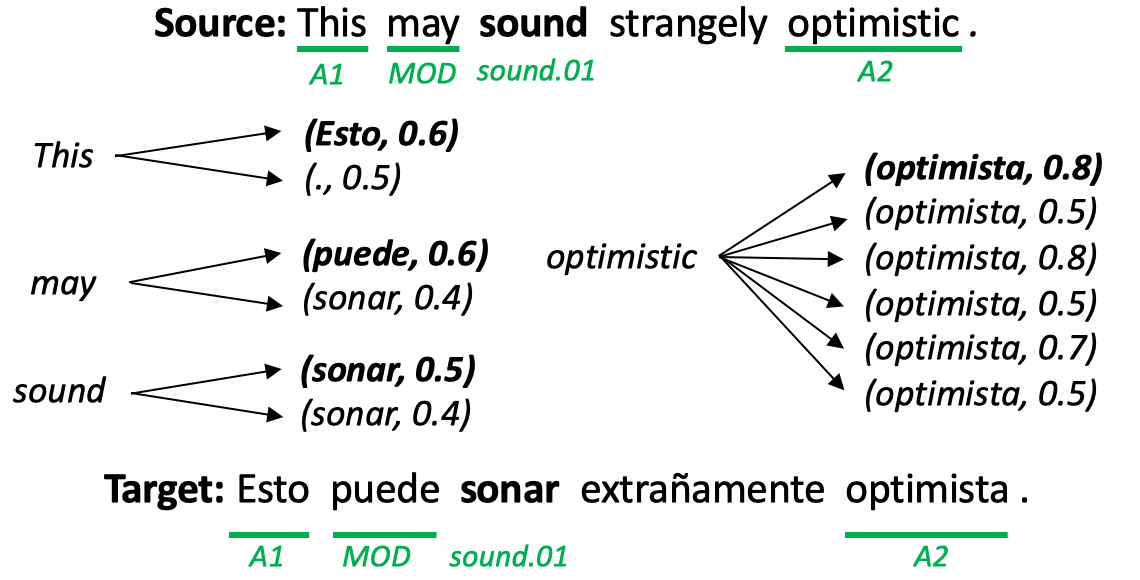}
  \caption{We map word pieces to full-words and apply filters to obtain final source-to-target word alignments.}
\label{fig:alignments_tree}
\end{figure}

\textbf{Word Alignments.} For each column in $SM$, we choose the \textit{k} most similar pairs ($v_{s}$, $v_{t}$) \footnote{$k$ is a hyperparameter which we chose by hand. The best results were obtained with $k$=2.}. This is analogous to a $\mathcal{A}_{S'\rightarrow T'}$ alignment \footnote{Conversely, we can simulate a $\mathcal{A}_{T'\rightarrow S'}$ alignment by defining a similar process for each row in the matrix.}. The alignment is done from full-word $w_s$ to  full-word $w_t$, meaning that for each $v_s$, instead of adding a $v_s \rightarrow v_t$ alignment, we retrieve the full-word $w_s$ to which $v_s$ belongs and the $w_t$ to which $v_t$ belongs and add a $w_s \rightarrow w_t$ alignment to the list of candidates for $w_s$. At this step, we still permit one-to-many mappings, which means that a $w_s$ can be associated to more than one $w_t$ candidates. We retain a dictionary $D=\{w_s: [(w_{t_1}, sim_{t_1}) ... (w_{t_x} , sim_{t_x})] | w_s \epsilon S \}$ with their associated similarity scores to keep track of the candidates. See Figure \ref{fig:alignments_tree} for an example.

\textbf{Alignment Modes.}
When projecting annotations to the translated training sections, we are confronted with the same \textit{special cases} that we identified in the test set. In the absence of human validation, we have to define filters to eliminate noisy alignments. By only keeping the intersection of alignments $\mathcal{A}_{S\rightarrow T} \bigcap \mathcal{A}_{T\rightarrow S}$, we can get rid of a considerable amount of noisy alignments, however this comes at the cost of a very low recall and a sparsely labeled dataset. Since, we are using an accurate word-similarity measure instead of (noisier) word alignments, we can encourage higher recall by considering all $\mathcal{A}_{S\rightarrow T}$ alignments and include additional filters to get rid of noisy labels and thus preserve high precision. In (\S \ref{sec:label_projection}) we describe in detail the experiments that support this assumption.

\begin{table*}[t!]
\centering
\resizebox{0.99\textwidth}{!}{

\begin{tabular}{lrrr|rrr|rrr|rrr}
\toprule
{ }      & \multicolumn{3}{c}{{ EN}}                                                 & \multicolumn{3}{c}{DE}                                                     & \multicolumn{3}{c}{ES}                                                     & \multicolumn{3}{c}{FR}                                                     \\
{ \textbf{X-SRL}} & { Sents}  & { Preds}  & { Args}    & Sents  & { Preds} & { Args}    & Sents  & { Preds} & { Args}    & Sents  & { Preds} & { Args}    \\ %\midrule
\cmidrule(lr){2-4}  \cmidrule(lr){5-7} \cmidrule(lr){8-10} \cmidrule(lr){11-13}
{Train} & { 39,279} & { 92,908} & { 238,887} & 39,279 & { 60,861}     & { 134,714} & 39,279 & { 68,844}     & { 154,536} & 39,279 & { 67,878}     & { 154,279} \\
{Dev}   & { 1,334}  & { 3,321}  & { 8,407}   & 1,334  & { 2,152}      & { 4,584}   & 1,334  & { 2,400}      & { 5,281}   & 1,334  & { 2,408}      & { 5,388}   \\
Test                         & 2,399                         & 5,217                         & 14,156                         & 2,213  & 4,086                              & 11,050                         & 2,346  & 4,376                              & 10,529                         & 2,095  & 3,770                             & 9,854   \\      \bottomrule
\end{tabular}

}
\caption{Overall statistics for X-SRL.} %Statistics: For each language we show nb.\ of sentences,  predicates and arguments in \textit{train/dev/test}.}
\label{tab:X-SRL_Data}
\end{table*}

\textbf{Filtered Projection.}
\footnote{\url{https://github.com/Heidelberg-NLP/xsrl_mbert_aligner}} 
%The first filter is to only look  at the source words of interest: 
First of all, we eliminate a considerable amount of potential noise by only looking at the $w_s$'s that hold a predicate or argument label, while ignoring the rest. Next, for each labeled source predicate, we retrieve from $D$ the list of target candidates and keep only those that bear a verbal POS tag.
If the list contains more than one target candidate we keep the one with the highest score, and if the list is empty we do not project the predicate, as it will most likely instantiate a translation shift or nominalization. Light verbs should be automatically filtered with this method, since the alignment links a verb to a noun and is therefore dropped. For the case of arguments, we also retrieve the candidates from D. In the ideal case, all candidates belong to the same $w_t$ and we project the label to that word. Otherwise, we take the $w_t$ with more \textit{votes}, i.e. the $w_t$ that was added most often to the list of candidates. In case of a tie, we turn to the similarity score and transfer the argument label to the $w_t$ with the highest similarity\footnote{Score aggregation would be a straightforward way of computing similarities. However, \citet{Zhang20-bertScore} mention that while cosine similarity is good to rank semantic similarity, the computed magnitude is not necessarily proportional, therefore it is not a strict metric. For this reason, we only rely on scores as a decision factor in case of ties.}.

\section{Experiments and Evaluations}
\label{exps}

\subsection{Label Projection}
\label{sec:label_projection}

% Test Set comparison
\textbf{Intrinsic Evaluation.} Since our test sets are human-validated, we can use them to measure the quality of the label projection methods we have at hand. First, we test the effectiveness of our full method (mBERT+Filters) by comparing it to vanilla cosine similarity (mBERT only) as a projection tool. We apply each method to the test sentences and evaluate the automatically assigned labels against the gold labels provided by annotators.  We also show the performance differences when keeping all source to target alignments (S2T) vs.\ using the intersection of alignments (INTER) when projecting both predicates and arguments. In Table \ref{tab:s2t-intersect} the four combinations can be observed with their specific trade-offs. When using only mBERT with S2T alignments we have high recall but a very mediocre precision; when using INTER alignments we see big gains in precision at the expense of lower recall, as expected. On the other hand, mBERT+Filters obtains consistently better F1, with INTER showing similar behavior to what we observe with the vanilla method, yet with much better precision; however, using full S2T alignments \textit{with filters} gives us the best trade-off: we still achieve around 90\% precision and much better recall compared to INTER. This confirms that using S2T alignments (established using mBERT-based cosine similarity) combined with our filters are the best option for projecting labels.

% Train Set Density
\textbf{Extrinsic Evaluation.} Having settled our best method, we compare it with an SRL label projection software: ZAP \cite{akbik18-zap} \footnote{\url{www.github.com/zalandoresearch/zap}}, which also works with the three target languages studied in this paper. ZAP is a pipeline model that takes as input parallel $(\mathbf{S}, \mathbf{T})$ sentences, uses source syntactic and semantic parsers to obtain the annotations, and through a trained heuristic word alignment module that uses pre-computed word translation probabilities, it transfers the labels only when it considers the alignments to be valid, preferring to have fewer, but higher-quality annotations on the target side.

\begin{table}[]
\centering
\resizebox{0.48\textwidth}{!}{

\begin{tabular}{@{}llrrrrrr@{}}
\toprule
Method                        & Lang  & \multicolumn{3}{c}{INTER} & \multicolumn{3}{c}{S2T} \\
                              &       & P         & R        & F1       & P     & R     & F1            \\ \hline
\multirow{3}{*}{mBERT Only}   & EN-DE & 86.6      & 49.6     & 63.0     & 69.0  & 76.1  & 72.4          \\
                              & EN-ES & 83.8      & 68.2     & 75.2     & 70.0  & 84.8  & 76.7          \\
                              & EN-FR & 82.7       & 61.8      & 70.7      & 67.7   & 79.5   & 73.1           \\ \hline
\multirow{3}{*}{mBERT+Filters} & EN-DE & 96.1      & 51.8     & 67.4     & 92.5  & 65.8  & \textbf{76.9} \\
                              & EN-ES & 94.0      & 68.8     & 79.4     & 91.9  & 80.7  & \textbf{85.9} \\
                              & EN-FR & 91.7       & 63.7      & 75.2      & 88.9   & 74.8   & \textbf{81.2}  \\ 
\bottomrule
\end{tabular}

}
\caption{Examining different projection methods on our \textit{human-validated test set}: a) vanilla mBERT cosim (mBERT-Only) vs.\ adding filters (mBERT+Filters); b) INTER using intersective alignments vs.\ S2T using full source-to-target alignments. Using S2T alignments and applying filters yield highest F1 alignment score. }
\label{tab:s2t-intersect}

\end{table}

\begin{figure}
\centering
  \includegraphics[width=0.42\textwidth]{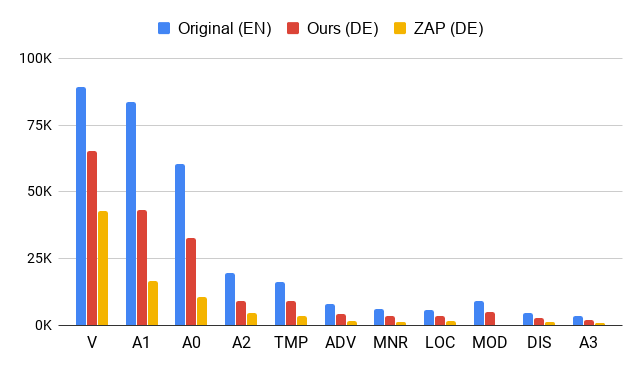}
  \caption{Ten most frequent labels obtained with two
  label projection methods: OURS vs.\ ZAP - on the German train set, compared to English source annotations.
  }
\label{fig:label_density}
\end{figure}

To compare our method to this baseline, we measure the density of the labels on the target training sets after applying both methods to project the labels\footnote{We consider the gold source labels for both methods, thus comparing only their projection performance}. Figure \ref{fig:label_density} shows the case of EN projected to DE where our method consistently recovers more labels from the source, resulting in a more densely annotated training set with comparable label distribution to the EN source. This trend is similar for Spanish and French (overall coverage relative to EN is: DE: 58.9\%, ES: 67.3\%,
FR: 66.9\%).

\begin{table}[t!]
\centering
\resizebox{0.48\textwidth}{!}{

\begin{tabular}{@{}lcccccccccccc@{}}
\toprule
                           & \multicolumn{6}{c}{\textbf{ZAP}}                                                                                           & \multicolumn{6}{c}{\textbf{OURS}}                                                                   \\
                           \cmidrule(lr){2-7} \cmidrule{8-13}
                           & \multicolumn{3}{c}{\textbf{PREDICATE}}                       & \multicolumn{3}{c}{\textbf{ARGUMENT}}                       & \multicolumn{3}{c}{\textbf{PREDICATE}}                     & \multicolumn{3}{c}{\textbf{ARGUMENT}} \\ %\cmidrule(l){2-13} 
\multicolumn{1}{l}{}      & \textbf{P} & \textbf{R} & \multicolumn{1}{c}{\textbf{F1}}   & \textbf{P} & \textbf{R} & \multicolumn{1}{c|}{\textbf{F1}}   & \textbf{P} & \textbf{R} & \multicolumn{1}{c}{\textbf{F1}} & \textbf{P} & \textbf{R} & \textbf{F1} \\ \midrule
\multicolumn{1}{@{}l|}{EN-DE} & 68.9       & 15.9       & \multicolumn{1}{c|}{25.9} & 72.7       & 15.6       & \multicolumn{1}{c|}{25.7} & 95.7       & 76.2       & \multicolumn{1}{c|}{\textbf{84.9}}        & 91.3       & 61.6       & \textbf{73.6}        \\
\multicolumn{1}{@{}l}{EN-ES} & 78.9       & 34.7       & \multicolumn{1}{c|}{48.2} & 68.7       & 30.5       & \multicolumn{1}{c|}{42.2} & 98.0       & 89.3       & \multicolumn{1}{c|}{\textbf{93.4}}        & 89.0       & 76.4       & \textbf{82.2}        \\
\multicolumn{1}{@{}l}{EN-FR} & 66.2       & 21.1       & \multicolumn{1}{c|}{32.0} & 66.5       & 24.4       & \multicolumn{1}{c|}{35.7} & 97.3       & 85.4       & \multicolumn{1}{c|}{\textbf{91.0}}        & 88.9       & 69.8       & \textbf{78.2}       \\
\bottomrule
\end{tabular}

}
\caption{We compare our best projection method with ZAP, a SOTA system for SRL label projection on our test sets. The recall of ZAP is extremely low, damaging their overall scores. In contrast, our method is very good at projecting verbal predicates and arguments.}
\label{LabelProjScore}
\end{table}

To investigate more deeply why ZAP performs so poorly compared to our method, we use the test sets to measure performance. We first evaluate the capacity to transfer source predicates to the target side. Table \ref{LabelProjScore} clearly shows that ZAP fails to transfer many predicates, perhaps because it has unreliable (or no) word-alignment probabilities for infrequent predicates and it is not fine-tuned for this domain (it was trained on Europarl). As a result, also the argument scores are very low, since for each predicate it misses, the system cannot recover any arguments. This highlights the main advantages of our method: by relying on a big multilingual language model i) we obtain high-quality word alignments featuring high precision \textit{and} recall, and ii) we do not need to re-train for other language pairs nor different domains.

\subsection{Training SRL Systems on X-SRL}

At this point we have attested the quality of the automatic method for creating the training sets.  Now, as an extrinsic evaluation, we will measure how well can different models learn from our data. To train the models we follow \cite{Zhou15-SRL,He17-SRL} in the sense that we feed the predicate in training and inference, and we process each sentence as many times as it has predicates, labeling one predicate-argument structure at a time. 

\textbf{mBERT fine-tuning.} In all settings, we fine-tune mBERT\footnote{We use BertForTokenClassification from \url{https://huggingface.co/transformers/}}. We use batch size of 16, learning rate of $5e^{-5}$ and optimize using Adam with weight decay \cite{loshchilov-18-AdamW} and linear schedule with warmup. We train for 5 epochs on our data and pick the epoch that performs best on \textit{dev}. Concretely, we explore three settings: The obvious baseline is i) to use only the available English high-quality labels for fine-tuning mBERT and apply zero-shot inference on the other three languages (we call this \textit{EN-tuned}). The other two settings are ii) to fine-tune each language independently with its respective training set (\textit{Mono}) and iii) using all the available data from the four languages to train a single model (\textit{Multi}). Table \ref{tab:BERT-SRL} shows that, as expected, for the \textit{EN-tuned} baseline, English reaches an F1 score of 91, and the other three languages can make good use of mBERT's knowledge in the zero-shot setting, reaching scores around 70. We also see that our training sets are more complete, obtaining, across the board, higher F1 scores than the training sets projected using ZAP. We observe that training on monolingual data results in improvements for all languages, and finally, the best setting is to use all data at once, improving the already robust mBERT results, and reaching scores of 77, 92, 81 and 78 for DE, EN, ES, FR respectively, about 8 points higher than the zero-shot baseline in the case of German.
\begin{table}[t!]
\centering
\resizebox{0.48\textwidth}{!}{

\begin{tabular}{@{}lrrrrrrrr@{}}
\toprule
\multicolumn{1}{c}{}  & \multicolumn{2}{c}{EN} & \multicolumn{2}{c}{DE}                   & \multicolumn{2}{c}{ES} & \multicolumn{2}{c}{FR} \\ 
\cmidrule(lr){2-3} \cmidrule(lr){4-5} \cmidrule(lr){6-7} \cmidrule(lr){8-9}
MODEL                  & ZAP    & OURS           & ZAP  & 
\multicolumn{1}{l}{OURS}          & ZAP   & OURS           & ZAP    & OURS           \\
 \cmidrule(lr){2-3} \cmidrule(lr){4-5} \cmidrule(lr){6-7} \cmidrule(lr){8-9}
mBERT EN-tuned           & 91.0    & 91.0           & 69.5 & \multicolumn{1}{l}{69.5}          & 75.1   & 75.1           & 71.9    & 71.9           \\
mBERT Mono (finetune)  & 91.0   & 91.0           & 58.6 & \multicolumn{1}{l}{76.1}          & 64.5  & 80.5           & 59.5   & 77.4           \\
mBERT Multi (finetune) & 92.4   & \textbf{92.9}  & 63.7 & \multicolumn{1}{l}{\textbf{77.0}} & 67.4  & \textbf{81.1}  & 64.1   & \textbf{78.3} \\
\bottomrule
\end{tabular}

}
\caption{F1 Score with Fine-tuning mBERT on our training data, created using ZAP vs.\ OUR projection method and evaluated on our test sets. We compare zero-shot (EN-tuned), mono- and multilingual settings.}
\label{tab:BERT-SRL}
\end{table}

\begin{table}[t!]
\centering
\resizebox{0.43\textwidth}{!}{

\begin{tabular}{@{}lllll@{}}
\toprule
MODEL             & EN   & DE   & ES   & FR   \\ \cmidrule{2-5}
\cite{daza-frank-2019-translate} Mono      &  90.9 & 67.6 & 56.2 & 58.1 \\
\cite{daza-frank-2019-translate} Multi     & 87.6 & 72.5 & 77.1 & 75.2 \\
\cite{Cai18-SRL} Mono & 91.4 & 76.5 & \textbf{82.6} & 80.3 \\
\cite{he-19-syntax} Mono    & \textbf{92.4} & 75.8 & 82.3 & 79.3 \\
\cite{he-19-syntax} Multi   & 92.1 & \textbf{77.3} & 82.5 & \textbf{80.4}\\
\bottomrule
\end{tabular}

}
\caption{F1 Score when training existing SRL models with our data and evaluating on our test. We compare monolingual (Mono) vs using all data available (Multi).}
\label{tab:SOTA-SRL}
\end{table}

\textbf{SOTA Models.} Next, we choose three SRL systems that show SOTA results on CoNLL-09 and train them using our data instead. Note that our results are not comparable since our train and test sets are completely different for ES and DE; also the EN results are not comparable since we only label verbal predicates; finally, FR is not present in CoNLL-09. Table \ref{tab:SOTA-SRL} summarizes the results. The model of \citet{daza-frank-2019-translate} is an Encoder-Decoder model that was designed for multilingual SRL. It performs poorly when trained on monolingual data but improves significantly when trained with more data (multilingual setting). The model of \citet{Cai18-SRL} adapts the biaffine attention scorer of \citet{DozatM-17-Biaffine} to the SRL task; we note that this model is not designed for handling multilingual data, therefore we only show the monolingual results, which still achieve the best score (82.6) for ES on our test data. Finally, \citet{he-19-syntax} generalizes and enhances the biaffine attention scorer with language-specific rules that prune arguments to achieve SOTA on all languages in CoNLL-09. When training this model using our data it achieves the highest scores for EN in the \textit{Mono} setting and for DE and FR when trained with multilingual data. In sum, using our new corpus to train multilingual SRL systems, with SOTA models and fine\-tu\-ning mBERT, we find evidence that the models can use the multilingual annotations for improved performance, especially for the weaker languages.

%\bluenote{Results show that training models with multilingual data results beneficial especially for the weaker languages, confirming our intuition that a multilingual dataset is helpful.}

%that the corpus design and the quality and density of annotations allows the models to generalize over language.}
\section{Conclusions}
\label{conclusions}

In this paper, we present the first fully parallel SRL dataset with homogeneous annotations for four different languages. We included human-validated test sets where we address the linguistic difficulties that emerge when transferring labels across languages -- despite transferring gold labels from the source. We introduce and evaluate a novel effective and portable automatic method to transfer SRL labels that relies on the robustness of Machine Translation and multilingual BERT and therefore could be straightforwardly applied to produce SRL data in other languages. Finally, we included an extrinsic evaluation where we train SRL models using our data and obtain consistent results that showcase the generalization capacities emerging from our new 4-way multilingual dataset. Future work should address the application of our method to more and typologically more divergent languages.

\section*{Acknowledgements}
We thank our reviewers for their insightful comments. This research was funded by the Leibniz ScienceCampus Empirical Linguistics
and Computational Language Modeling, supported by Leibniz Association (grant no. SAS2015-IDS-LWC) and by the Ministry of Science, Research, and Art of Baden-\-Wurttemberg. We thank NVIDIA Corporation for a donation of GPUs used in this research.

% \section*{Acknowledgments}

% The acknowledgments should go immediately before the references. Do not number the acknowledgments section.
% Do not include this section when submitting your paper for review.

\bibliographystyle{acl_natbib}
\bibliography{emnlp2020}

\begin{thebibliography}{35}
\expandafter\ifx\csname natexlab\endcsname\relax\def\natexlab#1{#1}\fi

\bibitem[{Akbik et~al.(2015)Akbik, Chiticariu, Danilevsky, Li, Vaithyanathan,
  and Zhu}]{Akbik15-Proj}
Alan Akbik, L.b Chiticariu, M.b Danilevsky, Y.b Li, S.b Vaithyanathan, and H.b
  Zhu. 2015.
\newblock \href
  {http://www.scopus.com/inward/record.url?eid=2-s2.0-84943776048&partnerID=40&md5=cd7425e611434e0a22c1790e0657ec01}
  {{Generating high quality proposition banks for multilingual semantic role
  labeling}}.
\newblock \emph{ACL-IJCNLP 2015}, 1:397--418.

\bibitem[{Akbik and Li(2016)}]{Akbik-PolyglotSRL}
Alan Akbik and Yunyao Li. 2016.
\newblock \href {https://doi.org/10.18653/v1/P16-4001} {{POLYGLOT:}
  multilingual semantic role labeling with unified labels}.
\newblock In \emph{Proceedings of {ACL-2016} System Demonstrations, Berlin,
  Germany, August 7-12, 2016}, pages 1--6. Association for Computational
  Linguistics.

\bibitem[{Akbik and Vollgraf(2018)}]{akbik18-zap}
Alan Akbik and Roland Vollgraf. 2018.
\newblock \href {https://www.aclweb.org/anthology/L18-1344} {{ZAP}: An
  open-source multilingual annotation projection framework}.
\newblock In \emph{Proceedings of the 11th Language Resources and Evaluation
  Conference}, Miyazaki, Japan. European Language Resource Association.

\bibitem[{Aminian et~al.(2019)Aminian, Rasooli, and Diab}]{AminianRD19-RawSRL}
Maryam Aminian, Mohammad~Sadegh Rasooli, and Mona~T. Diab. 2019.
\newblock \href {https://doi.org/10.18653/v1/w19-0417} {Cross-lingual transfer
  of semantic roles: From raw text to semantic roles}.
\newblock In \emph{Proceedings of the 13th International Conference on
  Computational Semantics, {IWCS} 2019, Long Papers, Gothenburg, Sweden, May
  23-27 May, 2019}, pages 200--210. Association for Computational Linguistics.

\bibitem[{Baker et~al.(1998)Baker, Fillmore, Lowe, Baker, Fillmore, and
  Lowe}]{Baker98-SRL}
Collin~F. Baker, Charles~J. Fillmore, John~B. Lowe, Collin~F. Baker, Charles~J.
  Fillmore, and John~B. Lowe. 1998.
\newblock {The Berkeley} {FrameNet} {Project}.
\newblock In \emph{Proceedings of the 17th international conference on
  Computational linguistics}, volume~1, page~86, Morristown, NJ, USA.
  Association for Computational Linguistics.

\bibitem[{Bonial et~al.(2015)Bonial, Hwang, Bonn, Conger, Babko-Malaya, and
  Palmer}]{bonial-12-propbank}
Claire Bonial, Jena~D. Hwang, Julia Bonn, Kathryn Conger, Olga Babko-Malaya,
  and Martha Palmer. 2015.
\newblock :english {P}rop{B}ank annotation guidelines.
\newblock In \emph{Center for Computational Language and Education Research
  Institute of Cognitive Science University of Colorado at Boulder}.

\bibitem[{Cai et~al.(2018)Cai, He, Li, and Zhao}]{Cai18-SRL}
Jiaxun Cai, Shexia He, Zuchao Li, and Hai Zhao. 2018.
\newblock A full end-to-end semantic role labeler, syntactic-agnostic over
  syntactic-aware?
\newblock In \emph{{COLING}}, pages 2753--2765. Association for Computational
  Linguistics.

\bibitem[{Carreras and M\`{a}rquez(2005)}]{Carreras05-SRL}
Xavier Carreras and Llu\'{\i}s M\`{a}rquez. 2005.
\newblock \href {http://dl.acm.org/citation.cfm?id=1706543.1706571}
  {Introduction to the conll-2005 shared task: Semantic role labeling}.
\newblock In \emph{Proceedings of the Ninth Conference on Computational Natural
  Language Learning}, CONLL '05, pages 152--164, Stroudsburg, PA, USA.
  Association for Computational Linguistics.

\bibitem[{Daza and Frank(2019)}]{daza-frank-2019-translate}
Angel Daza and Anette Frank. 2019.
\newblock \href {https://doi.org/10.18653/v1/D19-1056} {Translate and label! an
  encoder-decoder approach for cross-lingual semantic role labeling}.
\newblock In \emph{Proceedings of the 2019 Conference on Empirical Methods in
  Natural Language Processing and the 9th International Joint Conference on
  Natural Language Processing (EMNLP-IJCNLP)}, pages 603--615, Hong Kong,
  China. Association for Computational Linguistics.

\bibitem[{Devlin et~al.(2019)Devlin, Chang, Lee, and
  Toutanova}]{Devlin18-bert-exps}
Jacob Devlin, Ming-Wei Chang, Kenton Lee, and Kristina Toutanova. 2019.
\newblock \href {https://doi.org/10.18653/v1/N19-1423} {{BERT}: Pre-training of
  deep bidirectional transformers for language understanding}.
\newblock In \emph{Proceedings of the 2019 Conference of the North {A}merican
  Chapter of the Association for Computational Linguistics: Human Language
  Technologies, Volume 1 (Long and Short Papers)}, pages 4171--4186,
  Minneapolis, Minnesota. Association for Computational Linguistics.

\bibitem[{Dozat and Manning(2017)}]{DozatM-17-Biaffine}
Timothy Dozat and Christopher~D. Manning. 2017.
\newblock \href {https://openreview.net/forum?id=Hk95PK9le} {Deep biaffine
  attention for neural dependency parsing}.
\newblock In \emph{5th International Conference on Learning Representations,
  {ICLR} 2017, Toulon, France, April 24-26, 2017, Conference Track
  Proceedings}. OpenReview.net.

\bibitem[{Eger et~al.(2018)Eger, Daxenberger, Stab, and
  Gurevych}]{eger-18-cross-AM}
Steffen Eger, Johannes Daxenberger, Christian Stab, and Iryna Gurevych. 2018.
\newblock \href {https://www.aclweb.org/anthology/C18-1071} {Cross-lingual
  argumentation mining: Machine translation (and a bit of projection) is all
  you need!}
\newblock In \emph{Proceedings of the 27th International Conference on
  Computational Linguistics}, pages 831--844, Santa Fe, New Mexico, USA.
  Association for Computational Linguistics.

\bibitem[{Haji\v{c} et~al.(2009)Haji\v{c}, Ciaramita, Johansson, Kawahara,
  Mart\'{\i}, M\`{a}rquez, Meyers, Nivre, Pad\'{o}, \v{S}t\v{e}p\'{a}nek,
  Stra\v{n}\'{a}k, Surdeanu, Xue, and Zhang}]{Hajic09-SRL}
Jan Haji\v{c}, Massimiliano Ciaramita, Richard Johansson, Daisuke Kawahara,
  Maria~Ant\`{o}nia Mart\'{\i}, Llu\'{\i}s M\`{a}rquez, Adam Meyers, Joakim
  Nivre, Sebastian Pad\'{o}, Jan \v{S}t\v{e}p\'{a}nek, Pavel Stra\v{n}\'{a}k,
  Mihai Surdeanu, Nianwen Xue, and Yi~Zhang. 2009.
\newblock \href {http://dl.acm.org/citation.cfm?id=1596409.1596411} {The
  conll-2009 shared task: Syntactic and semantic dependencies in multiple
  languages}.
\newblock In \emph{Proceedings of the Thirteenth Conference on Computational
  Natural Language Learning: Shared Task}, CoNLL '09, pages 1--18, Stroudsburg,
  PA, USA. Association for Computational Linguistics.

\bibitem[{He et~al.(2017)He, Lee, Lewis, and Zettlemoyer}]{He17-SRL}
Luheng He, Kenton Lee, Mike Lewis, and Luke Zettlemoyer. 2017.
\newblock Deep semantic role labeling: What works and what’s next.
\newblock In \emph{Proceedings of the Annual Meeting of the Association for
  Computational Linguistics}.

\bibitem[{He et~al.(2019)He, Li, and Zhao}]{he-19-syntax}
Shexia He, Zuchao Li, and Hai Zhao. 2019.
\newblock \href {https://doi.org/10.18653/v1/D19-1538} {Syntax-aware
  multilingual semantic role labeling}.
\newblock In \emph{Proceedings of the 2019 Conference on Empirical Methods in
  Natural Language Processing and the 9th International Joint Conference on
  Natural Language Processing (EMNLP-IJCNLP)}, pages 5350--5359, Hong Kong,
  China. Association for Computational Linguistics.

\bibitem[{Huang et~al.(2019)Huang, Sun, Qiu, and Huang}]{huang-19-glossbertWSD}
Luyao Huang, Chi Sun, Xipeng Qiu, and Xuanjing Huang. 2019.
\newblock \href {https://doi.org/10.18653/v1/D19-1355} {{G}loss{BERT}: {BERT}
  for word sense disambiguation with gloss knowledge}.
\newblock In \emph{Proceedings of the 2019 Conference on Empirical Methods in
  Natural Language Processing and the 9th International Joint Conference on
  Natural Language Processing (EMNLP-IJCNLP)}, pages 3509--3514, Hong Kong,
  China. Association for Computational Linguistics.

\bibitem[{Koehn(2005)}]{Koehn05-Euro}
Philipp Koehn. 2005.
\newblock \href
  {http://homepages.inf.ed.ac.uk/pkoehn/publications/europarl-mtsummit05.pdf}
  {{Europarl: A Parallel Corpus for Statistical Machine Translation}}.

\bibitem[{Loshchilov and Hutter(2019)}]{loshchilov-18-AdamW}
Ilya Loshchilov and Frank Hutter. 2019.
\newblock \href {https://openreview.net/forum?id=Bkg6RiCqY7} {Decoupled weight
  decay regularization}.
\newblock In \emph{International Conference on Learning Representations}.

\bibitem[{Loureiro and Jorge(2019)}]{loureiro-19-BERT-WSD}
Daniel Loureiro and Al{\'\i}pio Jorge. 2019.
\newblock \href {https://doi.org/10.18653/v1/P19-1569} {Language modelling
  makes sense: Propagating representations through {W}ord{N}et for
  full-coverage word sense disambiguation}.
\newblock In \emph{Proceedings of the 57th Annual Meeting of the Association
  for Computational Linguistics}, pages 5682--5691, Florence, Italy.
  Association for Computational Linguistics.

\bibitem[{Marcheggiani et~al.(2017)Marcheggiani, Frolov, and
  Titov}]{March17-SRL}
Diego Marcheggiani, Anton Frolov, and Ivan Titov. 2017.
\newblock \href {http://aclweb.org/anthology/K17-1041} {A simple and accurate
  syntax-agnostic neural model for dependency-based semantic role labeling}.
\newblock In \emph{Proceedings of the 21st Conference on Computational Natural
  Language Learning (CoNLL 2017)}, pages 411--420, Vancouver, Canada.
  Association for Computational Linguistics.

\bibitem[{de~Marneffe et~al.(2014)de~Marneffe, Dozat, Silveira, Haverinen,
  Ginter, Nivre, and Manning}]{de-marneffe-14-universalDep}
Marie-Catherine de~Marneffe, Timothy Dozat, Natalia Silveira, Katri Haverinen,
  Filip Ginter, Joakim Nivre, and Christopher~D. Manning. 2014.
\newblock \href
  {http://www.lrec-conf.org/proceedings/lrec2014/pdf/1062_Paper.pdf} {Universal
  {S}tanford dependencies: A cross-linguistic typology}.
\newblock In \emph{Proceedings of the Ninth International Conference on
  Language Resources and Evaluation ({LREC}'14)}, pages 4585--4592, Reykjavik,
  Iceland. European Language Resources Association (ELRA).

\bibitem[{Meyers et~al.(2004)Meyers, Reeves, Macleod, Szekely, Zielinska,
  Young, and Grishman}]{meyers-04-nombank}
Adam Meyers, Ruth Reeves, Catherine Macleod, Rachel Szekely, Veronika
  Zielinska, Brian Young, and Ralph Grishman. 2004.
\newblock \href {https://www.aclweb.org/anthology/W04-2705} {The {N}om{B}ank
  project: An interim report}.
\newblock In \emph{Proceedings of the Workshop Frontiers in Corpus Annotation
  at {HLT}-{NAACL} 2004}, pages 24--31, Boston, Massachusetts, USA. Association
  for Computational Linguistics.

\bibitem[{Mulcaire et~al.(2018)Mulcaire, Swayamdipta, and
  Smith}]{Mulcaire18-SRL}
Phoebe Mulcaire, Swabha Swayamdipta, and Noah~A. Smith. 2018.
\newblock \href {https://doi.org/10.18653/v1/P18-2106} {Polyglot semantic role
  labeling}.
\newblock In \emph{Proceedings of the 56th Annual Meeting of the Association
  for Computational Linguistics (Volume 2: Short Papers)}, pages 667--672,
  Melbourne, Australia. Association for Computational Linguistics.

\bibitem[{Pado(2007)}]{Pado-07}
Sebastian Pado. 2007.
\newblock \emph{Cross-lingual Annotation Projection Models for Semantic Role
  Labeling}.
\newblock Ph.D. thesis.

\bibitem[{Pad\'{o} and Lapata(2009)}]{Pado09-Proj}
Sebastian Pad\'{o} and Mirella Lapata. 2009.
\newblock \href {http://dl.acm.org/citation.cfm?id=1734953.1734960}
  {Cross-lingual annotation projection of semantic roles}.
\newblock \emph{J. Artif. Int. Res.}, 36(1):307--340.

\bibitem[{Palmer et~al.(2005)Palmer, Kingsbury, and Gildea}]{Palmer05-SRL}
Martha Palmer, Paul Kingsbury, and Daniel Gildea. 2005.
\newblock The proposition bank: An annotated corpus of semantic roles.
\newblock \emph{Computational Linguistics}, 31.

\bibitem[{van~der Plas et~al.(2011)van~der Plas, Merlo, and
  Henderson}]{vanDerPlas11-Proj}
Lonneke van~der Plas, Paola Merlo, and James Henderson. 2011.
\newblock Scaling up automatic cross-lingual semantic role annotation.
\newblock In \emph{{ACL} (Short Papers)}, pages 299--304. The Association for
  Computer Linguistics.

\bibitem[{van~der Plas et~al.(2010)van~der Plas, Samardzic, and
  Merlo}]{vdPlas10-Proj}
Lonneke van~der Plas, Tanja Samardzic, and Paola Merlo. 2010.
\newblock Cross-lingual validity of propbank in the manual annotation of
  french.
\newblock In \emph{Linguistic Annotation Workshop}, pages 113--117. Association
  for Computational Linguistics.

\bibitem[{Pradhan et~al.(2012)Pradhan, Moschitti, Xue, Uryupina, and
  Zhang}]{PradhanCoNLL2012}
Sameer Pradhan, Alessandro Moschitti, Nianwen Xue, Olga Uryupina, and Yuchen
  Zhang. 2012.
\newblock \href {http://dl.acm.org/citation.cfm?id=2391181.2391183} {Conll-2012
  shared task: Modeling multilingual unrestricted coreference in ontonotes}.
\newblock In \emph{Joint Conference on EMNLP and CoNLL - Shared Task}, CoNLL
  '12, pages 1--40, Stroudsburg, PA, USA. Association for Computational
  Linguistics.

\bibitem[{Sennrich et~al.(2016)Sennrich, Haddow, and
  Birch}]{sennrich-etal-2016-improving}
Rico Sennrich, Barry Haddow, and Alexandra Birch. 2016.
\newblock \href {https://doi.org/10.18653/v1/P16-1009} {Improving neural
  machine translation models with monolingual data}.
\newblock In \emph{Proceedings of the 54th Annual Meeting of the Association
  for Computational Linguistics (Volume 1: Long Papers)}, pages 86--96, Berlin,
  Germany. Association for Computational Linguistics.

\bibitem[{Tiedemann and Agic(2016)}]{Tiedemann-16-AnnoMT}
J{\"{o}}rg Tiedemann and Zeljko Agic. 2016.
\newblock \href {https://doi.org/10.1613/jair.4785} {Synthetic treebanking for
  cross-lingual dependency parsing}.
\newblock \emph{J. Artif. Intell. Res.}, 55:209--248.

\bibitem[{Tyers et~al.(2018)Tyers, Sheyanova, Martynova, Stepachev, and
  Vinogorodskiy}]{tyers-18-AnnoMT}
Francis Tyers, Mariya Sheyanova, Aleksandra Martynova, Pavel Stepachev, and
  Konstantin Vinogorodskiy. 2018.
\newblock \href {https://doi.org/10.18653/v1/W18-6017} {Multi-source synthetic
  treebank creation for improved cross-lingual dependency parsing}.
\newblock In \emph{Proceedings of the Second Workshop on Universal Dependencies
  ({UDW} 2018)}, pages 144--150, Brussels, Belgium. Association for
  Computational Linguistics.

\bibitem[{Zhang et~al.(2020)Zhang, Kishore, Wu, Weinberger, and
  Artzi}]{Zhang20-bertScore}
Tianyi Zhang, Varsha Kishore, Felix Wu, Kilian~Q. Weinberger, and Yoav Artzi.
  2020.
\newblock \href {https://openreview.net/forum?id=SkeHuCVFDr} {{BERTScore:
  Evaluating Text Generation with {BERT}}}.
\newblock In \emph{8th International Conference on Learning Representations,
  {ICLR} 2020, Addis Ababa, Ethiopia, April 26-30, 2020}. OpenReview.net.

\bibitem[{Zhou and Xu(2015)}]{Zhou15-SRL}
Jie Zhou and Wei Xu. 2015.
\newblock End-to-end learning of semantic role labeling using recurrent neural
  networks.
\newblock In \emph{Proceedings of the 53rd Annual Meeting of the Association
  for Computational Linguistics and the 7th International Joint Conference on
  Natural Language Processing (Volume 1: Long Papers)}, pages 1127--1137,
  Beijing, China. Association for Computational Linguistics.

\bibitem[{Ziemski et~al.(2016)Ziemski, Junczys-Dowmunt, and
  Pouliquen}]{Ziemski16-UN}
Michał Ziemski, Marcin Junczys-Dowmunt, and Bruno Pouliquen. 2016.
\newblock The united nations parallel corpus v1.0.
\newblock In \emph{Proceedings of the Tenth International Conference on
  Language Resources and Evaluation (LREC 2016)}, Paris, France. European
  Language Resources Association (ELRA).

\end{thebibliography}

\end{document}